%% file: main.tex
\documentclass[11pt]{article}
\usepackage[dvipsnames]{xcolor}
\usepackage{times}
\usepackage{latexsym}
\usepackage{tikz-dependency}
\usepackage{amsmath}
\usepackage{setspace}
\usepackage{multirow}
\usepackage{float}
\usepackage{bbm}
\usepackage{pgfplots}
\usepackage{hyperref}
\usetikzlibrary {shadows,shapes.symbols}
\usepackage{graphicx} %
\usepackage{stfloats} %
\usepackage{makecell} %
\usepackage{subcaption}
\usepackage{lipsum}

\usepackage[]{ACL2023}

\usepackage{microtype}

\newcommand{\asstring}[1]{\textit{#1}}

\title{
Unsupervised Mapping of Arguments of Deverbal Nouns \\ to Their Corresponding Verbal Labels
} %

\author{Aviv Weinstein \\
  Department of Computer Science \\
  Bar-Ilan University \\
  \texttt{aviv.wn@gmail.com} \\\And
  Yoav Goldberg \\
  Department of Computer Science \\
  Bar-Ilan University \\
  \texttt{yoav.goldberg@gmail.com} \\}
\date{}

\begin{document}
    \maketitle
    
    \input{0_abstract}

    \section{Introduction}
    \label{sec:Introduction}

\input{1_introduction}

    \section{Deverbal Nouns}
    \label{sec:DeverbalNouns}
    \input{2_deverbal_nouns}

    \section{Related Works}
    \label{sec:RelatedWorks}
    \input{3_related_works}

    \section{Task Definition}
    \label{sec:TaskDefinition}
    \input{4_task_definition}

    \section{Methodology}
    \label{sec:Methodology}

\input{5_methodology}

    \section{Evaluation Data}
    \label{sec:EvaluationData}
    \input{6_evaluation_data}

    \section{Experiments and Results}
    \label{sec:ExperimentsAndResults}

\input{7_experiments_and_results}

    \section{Conclusions}
    \label{sec:Conclusions}
    \input{8_conclusions}

    \section*{Limitations}
    \label{sec:Limitations}
    \input{limitations}

    \section*{Ethics Statement} Like all works that depend on embeddings, the resulting models may be biased in various ways. Users should take this into consideration when deploying them in products.

    \section*{Acknowledgements} This project has received funding from the European Research Council (ERC) under the European Union's Horizon 2020 research and innovation programme, grant agreement No. 802774 (iEXTRACT).

    \bibliography{references}
    \bibliographystyle{acl_natbib}

    \clearpage

    \appendix
    \input{appendix.tex}

\end{document}

%% file: 0_abstract.tex
\begin{abstract}
    
Deverbal nouns are nominal forms of verbs commonly used in written English texts to describe events or actions, as well as their arguments. However, many NLP systems, and in particular pattern-based ones, neglect to handle such nominalized constructions. The solutions that do exist for handling arguments of nominalized constructions are based on semantic annotation and require semantic ontologies, making their applications restricted to a small set of nouns. We propose to adopt instead a more syntactic approach, which maps the arguments of deverbal nouns to the universal-dependency relations of the corresponding verbal construction. We present an unsupervised mechanism---based on contextualized word representations---which allows to enrich universal-dependency trees with dependency arcs denoting arguments of deverbal nouns, using the same labels as the corresponding verbal cases. By sharing the same label set as in the verbal case, patterns that were developed for verbs can be applied without modification but with high accuracy also to the nominal constructions.

\end{abstract}

%% file: 1_introduction.tex
\begin{figure}[h]
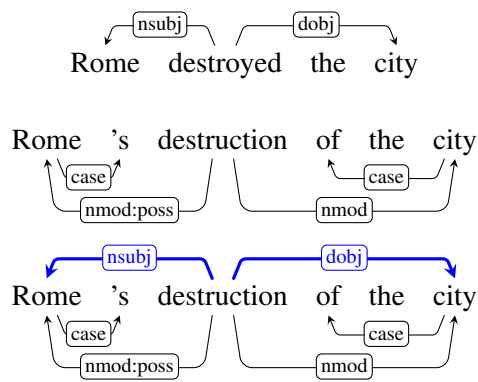

    \begin{gather*}
    \scalebox{1}{
        \begin{dependency}[theme = default]
            \begin{deptext}[column sep=0.2cm]
                Rome \& destroyed \& the \& city \\
            \end{deptext}
            \depedge[edge above,edge height=1.5ex]{2}{1}{nsubj}{3ex}
            \depedge[edge above,edge height=1.5ex]{2}{4}{dobj}{3ex}
        \end{dependency}
    }
    \\
    \scalebox{1}{
        \begin{dependency}[theme = default]
            \begin{deptext}[column sep=0.2cm]
                Rome \& 's \& destruction \& of \& the \& city \\
            \end{deptext}
            \depedge[edge below,edge height=4ex]{3}{1}{nmod:poss}{3ex}
            \depedge[edge below,edge height=1.5ex]{1}{2}{case}{3ex}
            \depedge[edge below,edge height=4ex]{3}{6}{nmod}{3ex}
            \depedge[edge below,edge height=1.5ex]{6}{4}{case}{3ex}
        \end{dependency}
    }
    \\
    \scalebox{1}{
        \begin{dependency}[theme = default]
            \begin{deptext}[column sep=0.2cm]
                Rome \& 's \& destruction \& of \& the \& city \\
            \end{deptext}
            \depedge[edge below,edge height=4ex]{3}{1}{nmod:poss}{3ex}
            \depedge[edge below,edge height=1.5ex]{1}{2}{case}{3ex}
            \depedge[edge below,edge height=4ex]{3}{6}{nmod}{3ex}
            \depedge[edge below,edge height=1.5ex]{6}{4}{case}{3ex}
            \depedge[edge above,edge height=1.5ex,edge style={blue,very thick}, label style={text=blue, draw=blue}]{3}{1}{nsubj}
            \depedge[edge above,edge height=1.5ex,edge style={blue,very thick}, label style={text=blue, draw=blue}]{3}{6}{dobj}
        \end{dependency}
    }
    \end{gather*}
    \caption{Example of our task. Top: verbal argument structure. Middle: nominal argument structure. Bottom: nominal structure enriched with corresponding verbal argument labels (thick blue edges).}
    \label{verb_vs_noun_ud}
\end{figure}

Systems that aim to extract and summarize information from large text collections often revolve around the concept of predicates and their arguments. Such predicates are often realized as verbs (\asstring{the performers interpret the music}), but the same predicative concepts can also be realized as nouns (\asstring{musical interpretation by the performers}). This process of realizing verbal predicates as nouns is called \emph{nominalization}, and it involves changing the syntactic structures around the content words participating in the construction, while keeping its semantics the same. In this work, we are interested in mapping arguments of nominal constructions that appear in text, to the corresponding ones in verbal structures (i.e., to identify the syntactic object role of \emph{music} and syntactic subject role of \emph{performers}, in \asstring{music interpretation by the performers}).

\paragraph{Nominalizations,} also known as nominal predicates, are nouns derived from words of a different part of speech, such as verbs or adjectives. For example, in English\footnote{While this work focuses on English nominalizations, the phenomena itself is not English specific.}, the nominalization \asstring{interpretation} is derived from the verb \asstring{interpret}, and the nominalization \asstring{precision} is related to the adjective \asstring{precise}. The usage of nominalizations is widespread in English text, and according to \citet{gurevich2007deverbal}, about half of all sentences in written texts contain at least one nominalization. In our work, we observed a ratio of 120k nominalizations to 180k verbs, in a random collection of 100k Wikipedia sentences. Thus, interpretation of nominalizations is central to many language understanding tasks. In the current work, We focus on nominalizations which are derived solely from verbs, commonly called deverbal nouns.

\paragraph{Existing attempts} around identifying arguments of nominalizations
either rely on a predefined semantic roles ontology (e.g., SRL based roles such as those in VerbNet \cite{schuler2005verbnet} or FrameNet \cite{baker1998berkeley}) as suggested by \citet{pradhan2004parsing}, \citet{pado2008semantic} and \citet{zhao2020unsupervised}, or consider a limited subset of nominalized structures (\citet{lapata2000automatic} and \citet{gurevich-waterman-2009-mining}). Early works approached the task in a fully supervised manner (\citet{lapata2000automatic}, \citet{pradhan2004parsing}), hence suffering from insufficient annotated nominal data. To overcome that, \citet{pado2008semantic} and more recently \citet{zhao2020unsupervised} considered a transfer scenario from verbal arguments to nominal arguments while assuming only supervised data for verbs. Nevertheless, their methods were limited to specific predicates, even with extensive annotated verbal data. Moreover, the previous works considered each a different set of argument types due to supervision constraints.

\paragraph{Our Proposed Task} Rather than relying on a predefined semantic roles ontology, in this work we propose to map the arguments of deverbal nouns to the \emph{syntactic} arguments of the corresponding active verbal form. This allows us to define a task with a consistent and a restricted label set (syntactic subject, syntactic object, syntactic prepositional modifier with preposition X), while still maintaining expressivity: if one knows how to extract the verbal argument from the active verbal form, they will be able to also extract the nominal ones.

A natural formulation is to ask ``How will this verb arguments be realized in a deverbal noun construction?''. However, this approach is problematic, as the same verbal structure, e.g. \asstring{IBM appointed Sam as manager}, can be realized in many different ways around the same nominalization, including:
\asstring{IBM's appointment of Sam as manager}, \asstring{Sam's appointment as manger by IBM} and \asstring{Sam's IBM appointment as manager}.

One solution would be to ask for all the possible nominal realizations. This is the approach taken by nominalization lexicons such as NomLex \cite{macleod1998nomlex}. However, this is also problematic in practice, as the different possible syntactic structures may conflict when encountering a nominalization within a sentence (\asstring{IBM's appointment} vs. \asstring{Sam's appointment}).

We resolve this by asking the opposite question: ``given a nominalized instance within a sentence and its set of arguments, how will these arguments map to those of an active verb construction?''. That is, rather than asking ``how will this verbal construction be realized as a nominal one'' we ask ``how will this nominal case be realized as an active verb construction''. Using this formulation, we define a corpus enrichment task, in which we take in a corpus of syntactic trees, and annotate each deverbal noun case with its nominal arguments, using the corresponding verbal argument labels. An example of the trees enrichment is provided in Figure \ref{verb_vs_noun_ud}.

\paragraph{Potential Utility} Our motivation follows that of \citet{tiktinsky2020pybart}: we imagine the use of the enhanced trees in systems that integrates universal dependency trees \cite{nivre2016universal} as part of their logic, using machine-learned or pattern-based techniques. Our proposed enrichment will allow users to search for a verb construction, and retrieve also nominal realizations of the same relation.

One proposed usage case regards the task of Open Information Extraction \citep[OpenIE;][]{etzioni2008open}, which refers to the extraction of relation tuples from plain text,
without demanding a predefined schema. These tuples can be extracted from both verbal and nominal phrases, e.g., the tuple (Steve Jobs; founded; Apple) from the phrase \asstring{Steve Jobs founded Apple} and the tuple (IBM; research) from the phrase \asstring{IBM's research}. Some OpenIE systems, such as Renoun \cite{yahya2014renoun} and \citeposs{angeli2015leveraging} system, integrate rule-based patterns to extract such relations from nominal phrases, e.g., (X; Y) from phrases of the structure ``X's Y''. However, these patterns can be misleading, as \asstring{IBM's research} interprets differently from \asstring{Rome's destruction} (IBM researched vs. Rome was destructed), leading to contradicting relations. To overcome that, we suggest using verb-based patterns to extract relations from nominal phrases, upon integrating our enhanced trees. Concretely, based our enhanced trees, an OpenIE system can use a pattern that detects the nsubj-phrase and dobj-phrase for both verbs and nouns, to construct the relation tuple (nsubj; verb/noun; dobj). With this approach, different nominal phrases with the same syntactic structure, would properly map to different ordered relations, as (destruction; Rome) for the phrase \asstring{Rome's destruction}.

\paragraph{An Unsupervised Approach} We take an unsupervised approach to this nominal-to-verbal argument mapping, relying on pre-trained contextualized word representations. The intuition behind our approach is that in order to resolve nominal arguments to verbal ones, there are two prominent signals: the semantic types of the arguments, and their syntactic configuration with respect to their predicate. We hypothesize that pre-trained contextualized word embeddings capture both of these signals (as shown in Section \ref{sec:argument_labeling_exp}), and also capture the similarities between the verbal and nominal cases (as demonstrated in Appendix \ref{appendix:verb_noun_argument_similarity}). Briefly, our approach works by identifying the candidate arguments of each deverbal noun instance, retrieving a set of sentences containing the corresponding active verb form, encoding both the deverbal noun instance and the active verb sentences using a masked language model, and searching for a mapping that maximizes some similarity metric between the nominal argument candidates and the verbal instances.

\paragraph{Our contributions in this work} are thus two-fold: (1) we formulate the task of aligning nominal arguments to the arguments of their corresponding active verbal form; and (2) we propose an unsupervised method for tackling this task. We also provide code\footnote{Our code is available at \url{https://github.com/AvivWn/NounVerbUDTransfer}} for enriching universal dependency trees \cite{nivre2016universal} with nominal arguments.

%% file: 2_deverbal_nouns.tex
Deverbal nouns are one type of nominalizations which are derived specifically from verbs, e.g., the deverbal noun \asstring{treatment} is derived from the verb \asstring{treat}. The events represented by deverbal nouns are described using phrases in the sentence that complement the nouns. The arguments of the deverbal noun correspond to the arguments of the matching verb; each matches a different question about the action taken. For instance, in the phrase \asstring{professional treatment of illness}, \asstring{professional} refers to the actor/subject of the verb \asstring{treat} (professionals), and \asstring{illness} refers to the object of the action \asstring{treat}.

The deverbal nouns, as typical nouns, are most often complemented by other noun phrases (\asstring{treatment of illness}, \asstring{his treatment} and \asstring{health treatment}) and adjectives (\asstring{professional treatment}). Implicit and other types of complementing arguments are not considered part of this work's scope. Each deverbal noun defines a unique structure of these arguments, assigning different roles for the same typed arguments. For instance, consider the phrases \asstring{time preference of the individual} and \asstring{individual waste of time}, which match the same syntactic structure (``noun-compound \textit{of} noun''). However, the first sentence matches the structure ``Obj Noun of Subj'' (``individuals$_2$ prefer time$_1$''), and the second sentence refers to the structure ``Subj Noun of Obj'' (``individual$_1$ waste time$_2$''). Furthermore, even the same deverbal noun may demand different labels for similar arguments in different contexts. For example, in the phrase ``\asstring{Rome's destruction}'', \asstring{Rome} was destroyed, whereas in the phrase ``\asstring{Rome's destruction of the city}'', \asstring{Rome} is the destroyer. Therefore, the argument roles are not determined solely by syntactic structure, and incorporate a mix of syntactic configuration, argument semantics, and predicate-specific information.

%% file: 3_related_works.tex
Arguments of nominalizations were long investigated in the field of NLP. One early research explored the syntactic structure of the arguments and modeled the structure of many nominalizations, resulting in a detailed lexicon called NomLex \citep{macleod1998nomlex}. The lexicon seeks to describe the allowed complements structures for a nominalization and relate the nominal complements to the arguments of the corresponding verb. Following the publishing of NomLex, \citet{meyers1998using} described how an Information Extraction (IE) system could exploit the linguistic information in the NomLex lexicon. Yet, the suggested approach remained hardly utilized by further research, as many works only exploited the verb-noun pairs specified by the lexicon.

Regarding identifying and labeling nominalization's arguments, a supervised approach was suggested while considering various task settings. One preceding paper by \citet{lapata2000automatic} presented a probabilistic procedure to infer whether the modifier of a nominalization (the head noun) stands in subject or object relation with it. For instance, the algorithm should predict that the modifier's role in the phrase \asstring{child behavior} is subject since the phrase refers to the \asstring{child} as the agent of the action described by the verb \asstring{behave}. Stated differently, this procedure focuses on extracting only one specific argument of nominalizations in a noun phrase. Another distinguished paper by \citet{pradhan2004parsing} considered FrameNet-based \citep{baker1998berkeley} semantic arguments of nominalizations and applied a machine learning framework for eventive nominalizations in English and Chinese, aiming to identify and label their arguments. Finally, \citet{kilicoglu2010arguments} published a similar approach for nominalizations used in biomedical text.

Some related works acknowledge the shortage of labeled argument nominalizations and suggest unsupervised methods for data expansion based on labeled argument verbs. Similarly to ours, these works exploited the similarity and alignment of the noun-verb arguments. For example, \citet{pado2008semantic} and \citet{zhao2020unsupervised} considered the argument labeling task for nominalizations in a setup where the verbal sentences are human labeled, and with regards to semantic role labeling (SRL) arguments. \citet{pado2008semantic} exploited the similarities between the argument structure of event nominalizations and corresponding verbs while utilizing common syntactic features and distributional-semantic similarities. More recently, \citet{zhao2020unsupervised} suggested a variational auto-encoder method, in which the labeler serves as an encoder, whereas the decoder generates the selectional preferences of the arguments for the predicted roles.

A different approach taken by \citet{gurevich-waterman-2009-mining} using a fully unsupervised manner while automatically extracting and labeling verbal arguments of verbs from a large parsed corpus of Wikipedia. This approach resembles an intermediate stage of ours yet differs as it considers a reduced set of argument types (subject and object) and a reduced possible set of argument syntax for the nominalizations (possessive and `of' arguments). Lately, \citet{lee-etal-2021-paraphrasing} engaged with a different task with similar applications. They suggested an unsupervised method for paraphrasing clauses with nominalizations into active verbal clauses.

%% file: 4_task_definition.tex
As discussed in the introduction, we define a task of labeling the arguments of deverbal nouns within a sentence, with labels of the arguments in the corresponding active verb constructions. Here we provide a more complete and formal definition. While our aim is to label all of the deverbal nouns in a given corpus, here we focus on describing the task with relation to a single instance of a sentence and a deverbal noun within it.

We consider the syntactic arguments of active verbal forms to belong to the set $L$ consisting of the universal dependency relations \emph{nsubj}, \emph{dobj} and \emph{nmod:X}, where $X$ is a preposition (e.g., \emph{nmod:in}, \emph{nmod:on}, \emph{nmod:with}). In words, the syntactic subject, syntactic object, and arguments attached as prepositional phrases where the identity of the preposition is part of the relation. While these prepositions may correspond to many different semantic roles, for a given verb they usually indicate a concrete and unique role.

Formally, given a sentence with words $w_1,\ldots,w_n$, and a marked deverbal noun within the sentence (say in position $w_i$), we seek to find $K$ pairs of the form $(rel_k, w_{j_k})$, $1 \leq k \leq K$, where $rel_k \in \{nsubj, dobj, nmod\textit{:}X\}$ and $w_{j_k}$ is a word in the sentence ($j_k$ is an index of a sentence word). For simplicity, we also demand that every relation type cannot be repeated more than once in the identified set of pairs.
These pairs indicate arguments of the deverbal noun and their relations to it, expressed using an active-verb label set.

In Figure \ref{verb_vs_noun_ud}, the blue edges of the bottom tree indicate the output \emph{(nsubj, 1), (dobj, 6)}. Note that the task includes both the \emph{identification} of the arguments and their \emph{label assignment}.

%% file: 5_methodology.tex
\begin{figure*}[h]
    \begin{subequations}
        \begin{gather}
            \ell_n = \arg\max_\ell sim(\mathbf{a_n}, avg(\{ \mathbf{\tilde{a}} \mid \ell(\tilde{a}) = \ell, \tilde{a} \in \tilde{A} \})) \label{eq:neareset-avg-argument}
            \\
            \ell_n = \arg\max_\ell sum(\{ sim(\mathbf{a_n}, \mathbf{\tilde{a}}) \mid \ell(\tilde{a}) = \ell, \mathbf{\tilde{a}} \in knn(\mathbf{a_n}, \mathbf{\tilde{A}}, k) \}) \label{eq:k-nearest-argument}
        \end{gather}
    \end{subequations}
\end{figure*}

While we intend to handle all deverbal nouns in a given collection of sentences, here we focus on how to resolve a single deverbal noun. We identify deverbal nouns and their corresponding verbal forms based on a given lexicon of verb-noun pairs, which we consider as input. In this work, we use the NomLex lexicon \cite{macleod1998nomlex}, where future work can also replace this with a learned model. %

Given a deverbal noun within a sentence, we first identify its potential arguments. This is realized by searching a set of syntactic relations in the corresponding universal dependency tree (we use the UDv1 parser trained by \citet{tiktinsky2020pybart} via the spaCy toolkit\footnote{\url{https://spacy.io}}).
We then label the arguments by comparing their contextualized word embeddings to those of the corresponding verb arguments, in a set of sentences containing this verb (we further motivate this comparison in Appendix \ref{appendix:verb_noun_argument_similarity}). Finally, based upon the labeled arguments, we construct the final output as pairs of the arguments' label (i.e. verbal UD relation) and the arguments' head word.

\subsection{Argument Identification}

Given a sentence and a specific deverbal noun within, we first identify the phrases which could correspond to the desired arguments of the matching verb. The identified set of phrases %
is referred to as ``argument candidates''.  Naively, every phrase in the sentence can complement the deverbal noun and be considered as an argument, thus resulting in a relatively large set of candidates. To reduce this set, we consider the syntactic dependency tree of the sentence, searching for words that stand with direct dependency relation with the deverbal noun. Then, for every identified word we construct the argument candidate as the phrase corresponding to the subtree headed by this word according to the dependency tree. More specifically, we observed that arguments of deverbal nouns are realized using words that stand with the deverbal nouns in a small set of possible syntactic relations: \emph{nmod:poss}, \emph{compound}, \emph{amod}, and \emph{nmod:X}. Table \ref{arguments_ud_relations_table} provides an example of these syntactic relations, using argument candidates for the deverbal noun \asstring{analysis}. In Section \ref{sec:argument_identification_exp} we compare this approach and other considered approaches to identify the arguments.

\begin{figure}[h]
\begin{center}
    \begin{tabular}{c c} 
        \hline
        Phrase & UD Relation \\ [0.5ex] 
        \hline\hline
        \asstring{\textbf{his} analysis} & nmod:poss \\ 
        \asstring{\textbf{data} analysis} & compound \\
        \asstring{\textbf{linguistic} analysis} & amod \\
        \asstring{analysis of \textbf{the data}} & nmod:of \\
        \hline
    \end{tabular}
    \captionof{table}{The types of UD relations we used to identify candidate arguments, and their example with the deverbal noun \asstring{analysis}.}
    \label{arguments_ud_relations_table}
\end{center}
\end{figure}

\subsection{Argument Labeling}
\label{sec:argument_labeling}

Upon argument identification, we aim to label the identified argument candidates of the deverbal nouns, with the desired argument types (\emph{nsubj}, \emph{dobj}, \emph{nmod:X} or $\emptyset$), such that the labels align to the labels of the corresponding arguments in the active verbal form (the label $\emptyset$ indicates that this argument candidate is not in fact an argument of the noun, such as \emph{primary} in the phrase \emph{the primary influence}). For instance, in the sentence \asstring{The emperor's destruction of Paris}, we wish to label \asstring{the emperor} as \emph{nsubj} and \asstring{Paris} as \emph{dobj}, since the sentence can only be understood as the verbal sentence \asstring{The emperor destroyed Paris}.

Concretely, denote the argument candidates as $a_1, \ldots, a_N$. We need to assign them with labels $\ell_1, \ldots, \ell_N$, where $\ell_i \in \{\emptyset, nsubj, dobj, nmod\textit{:X}\}$, under the constraint that every two arguments $a_i$, $a_j$, can share labels if and only if they match the label $\emptyset$ (as emphasized in the defined task). %

We start from obtaining a set of verbal reference sentences $S$, containing $M$ sentences $s_1,\ldots,s_M$, each sentence $s_m$ contains the verbal form of the deverbal noun (these are obtained using a simple keyword search). In each of these instances $s_m$, we use simple active and passive verbal dependency patterns 
to identify the $A_m$ verbal arguments $\tilde{a}_1^m, ..., \tilde{a}_{A_m}^m$, labelled as $\tilde{\ell}_1^m, \ldots, \tilde{\ell}_{A_M}^m$. Intuitively, we now seek to find for each of our nominal argument $a_n$ the most similar verbal argument $\tilde{a}^m_j$, and match their labels. In our experiments, we obtained a set $S$ containing about 1,500 reference sentences\footnote{We considered $\ll$1,500 reference sentences for less frequent verbs.} regarding every verb that were required by the evaluation datasets.

We encode both the input sentence and the reference sentences using a contextualized encoder (we use BERT-large-uncased \cite{devlin2018bert} in this work), resulting in vectors $\mathbf{a_1},\ldots,\mathbf{a_N}$ for the input sentence and vectors $\mathbf{\tilde{a}_1^m}, ..., \mathbf{\tilde{a}_{A_m}^m}$ for each verb reference sentence $s_m$. We denote the entire set of verbal arguments as $\tilde{A}$ and the corresponding set of vectors as $\mathbf{\tilde{A}}$. We use a metric function $sim(\mathbf{a},\mathbf{\tilde{a}})$ over the pair of vectors to quantify their similarity (we use \emph{cosine} similarity in this work). We then choose the label of each nominal argument $a_n$ independently\footnote{We also experimented with jointly labeling several arguments, but did not see any benefit.} based on its closest neighbours in $\mathbf{\tilde{A}}$. We consider two variants: 
in the first one (\ref{eq:neareset-avg-argument}, nearest-avg-argument), we select the label $\ell_n$ by averaging the reference vectors for each verbal argument label, and then choosing the label whose corresponding average vector is the most similar to the nominal argument's vector.
In the second variant (\ref{eq:k-nearest-argument}, k-nearest-argument), we take the k-nearest verbal argument vectors (we use k=5) to the nominal argument vector. We compute the sum of similarities between $\mathbf{a_n}$ and each of the k-nearest vector $\mathbf{\tilde{a}}$ corresponding to each label, and choose the label with the highest sum.

For both labeling variants, we assign the label $\emptyset$ for arguments whose similarity with any other reference argument does not pass a chosen threshold.

%% file: 6_evaluation_data.tex
Our task is to identify arguments of deverbal nouns and assign each one of them a label from the set $L = \{nsubj, dobj, \textit{nmod:X}\}$. For evaluation, we need sentences with deverbal nouns whose arguments are labeled with these relations. For example, the deverbal noun \asstring{relocation} in the phrase \asstring{Family relocation to Manchester} should be labeled with the pairs \emph{(nsubj, 1)} and \emph{(nmod:to, 4)}, as specified in Section \ref{sec:TaskDefinition}.

We create three such evaluation datasets, the first based on a nominalization paraphrasing dataset, and the other two are based on the NomLex lexicon, while they differ by the coverage of deverbal nouns that they consider, as we further explain. Moreover, to compare our method's performance to earlier works, we consider the CoNLL-2009 dataset \cite{hajivc2009conll} for evaluation, as we discuss in \ref{sec:comparison_to_earlier_work}.

\paragraph{The paraphrasing-derived evaluation set} is derived from a manually annotated dataset for the task of paraphrasing sentences from nominal to verbal form \cite{lee-etal-2021-paraphrasing}. The original dataset includes a collection of 449 samples from 369 unique sentences representing 142 different verbs. Each sample represents a paraphrasing between the original nominalization phrase (from a given sentence) and a verbal clausal phrase, for instance \asstring{genetic analysis from a sample} which is paraphrased as \asstring{analyze genes from a sample}. For every paraphrasing sample, the dataset specifies the components of the nominal phrase within the structure ``\emph{adj/noun} nominalization \emph{prep} \emph{pobj}'', and the components of the active verbal phrase (``\emph{arg0} verb \emph{arg1} \emph{pp}'').

To construct our evaluation set based on this data, we first match each of the nominal components adj/noun and pobj with a verbal component from the set of arg0, arg1 and pp, choosing the one with the closest orthography to the nominal one. From this, we derive the verbal argument labeling for the components of the nominal phrase. 
Then, we replace each verbal label with its matching UD relation.\footnote{\emph{arg0} $\mapsto$ \emph{nsubj}, \emph{arg1} $\mapsto$ \emph{dobj}, \emph{pp} $\mapsto$ \emph{nmod:X}, where X is determined by the leading preposition.} Finally, for every nominal component we determine its head word position in the given context. The word positions paired with the matching verbal relations, construct a sample in our new paraphrasing-derived evaluation set. %

In the course of dataset construction, we filter out some data samples. To start with, data samples that specify two nominal components that match the same verbal component are removed from our dataset, as they do not fit the constraints of the defined task. For example, in the phrase \asstring{environmental assessment for the project} the combined components of the noun can be understood together as the object of the matching verb (\asstring{assess the environmental impact of the project}), hence resulting with two nominal arguments labeled with the same verbal relation. Secondly, we consider only the first single data sample for every repeated nominal phrase to ensure a single truth of labeling for every nominal phrase. Following the filtering process we remain with 309 samples with 122 different verbs.

\paragraph{The NomLex evaluation sets} are constructed using the NomLex lexicon.\footnote{We converted the NomLex lexicon from its original LISP-based formatting and phrase-structure trees, to a more modern form encoded in JSON and using UD syntactic relations. The code for this conversion is accessible at \url{https://github.com/AvivWn/NounVerbUDTransfer}.}
The NomLex lexicon contains a list of about 4k deverbal nouns, and for each of them specifies the various ways in which their arguments can be realized syntactically, and how they map to the corresponding verbal arguments. 
For example, an adapted NomLex entry for a deverbal noun like \asstring{destruction} would specify the related forms of the noun (i.e., the verb and other related deverbal nouns) and, most significantly, a set of dependency-tree patterns corresponding to several different realizations of the noun. Each dependency-tree pattern represents a set of labeled arguments in a specific dependency tree. For instance, the entry of \asstring{destruction} would contain a pattern that corresponds to the dependency structure shown in the middle of Figure \ref{verb_vs_noun_ud} and demands the labeling of \asstring{Rome} as subject and \asstring{city} as object. Hence, using a parsed dependency tree of a sentence with a deverbal noun, we can extract the labeled arguments in the sentence for any specified pattern that fulfills the sentence's dependency structure. However, this method does not allow for a definitive decision in many cases, as the lexicon often contains multiple labeled contradicting patterns. In Section \ref{sec:ExperimentsAndResults} we show that relying solely on NomLex results in a significantly lower precision.

We collect English Wikipedia sentences from \citet{guo2020wiki} that contain a deverbal noun, and for each sentence, we identify the deverbal noun's arguments and labels based on the adapted NomLex entry as described above. We discard sentences for which the entry suggests two or more different assignments, when matching two or more dependency patterns. We then map NomLex's labels into the corresponding dependency relations of the active verbal form. To match the examples in the paraphrasing dataset, we consider only data samples with two labeled arguments each. We divide the collected samples into two evaluation sets based on the verbal form of the represented deverbal nouns. $\textbf{NomLex}_{paraphrasing}$ considers only samples which refer to verbs that appear in the paraphrasing-derived corpus, whereas $\textbf{NomLex}_{other}$ considers samples that match 315 other verbs. In each evaluation set, we keep 25 labeled sentences for each verb.

\paragraph{Tune/Test Split} Our method is unsupervised but still requires tuning of hyperparameters. We keep a tuning subset for each origin of the evaluation set (paraphrasing-derived and NomLex), which is also used for evaluation during development. In the paraphrasing dataset, we sample 20\% of the dataset to construct the tuning set while keeping aside 80\% of the dataset for evaluation. Out of the 122 verbs in the paraphrasing-derived evaluation set, 12 appear only in the tuning set, 83 only in the test set, and 27 appear in both sets. The split aims to ensure that the results are not verb-specific and to prevent overfitting, as we do hyperparameter optimization on the tuning set, which does not contain all the verbs that appear in the test set. To tune the method for NomLex-based data, we perform a similar tune-test split on $\text{NomLex}_{paraphrasing}$ based upon the same tune-test verb division made for the paraphrasing evaluation set. Concretely, NomLex instances of the 12 tuning-only verbs and 83 test-only verbs are included only in the NomLex tuning set and test set, correspondingly; Instances of the 27 common verbs are divided into the tune-test sets in a 20\%-80\% ratio. Moreover, we preserve entirely $\text{NomLex}_{other}$ corpus for testing.

\paragraph{Evaluation Metrics}
We use two evaluation metrics:
\textbf{Relation-F1} is the F1 score of all the predicted word-relation pairs compared to the gold labeled pairs (without distinguishing argument labels, for comparability with \citet{zhao2020unsupervised} which uses CoNLL-2009 evaluation scorer \cite{hajivc2009conll}). \textbf{Exact-Match} scores how many noun instances had all their relations identified and labeled correctly. A predicted relation is considered correct if it matches both the same argument head word and the same label as the gold relation.

%% file: 7_experiments_and_results.tex
\begin{figure*}[hbt!]
\begin{center}
    \begin{tabular}{l c c c c c c}
    \hline
    \multicolumn{1}{c}{} & \multicolumn{2}{c}{Paraphrasing-derived} & \multicolumn{2}{c}{$\text{NomLex}_{paraphrasing}$} &
    \multicolumn{2}{c}{$\text{NomLex}_{other}$}\\
    \multicolumn{1}{c}{Method} & F1 & Exact & F1 & Exact & F1 & Exact \\
    \hline\hline
    baseline (NomLex-based)     & 43.42 & 7.66 & - & - & - & - \\
    all-subject                 & 27.67 & 0.00 & 37.04 & 0.00 & 41.52 & 0.00 \\
    all-object                  & 36.50 & 0.00 & 40.24 & 0.00 & 38.19 & 0.00 \\
    nearest-avg-argument        & 44.08 & 17.74 & 39.81 & 18.38 & 40.10 & 19.49 \\
    k-nearest-argument          & \textbf{62.93} & \textbf{36.29} & \textbf{53.74} & \textbf{34.98} & \textbf{53.67} & \textbf{35.06} \\
    \hline
    \end{tabular}

    \captionof{table}{The best results of the two suggested labelers on the three test sets, compared to the baseline process and the naive methods. Regarding metrics, `F1' refers to Relation-F1 and `Exact' refers to Exact-Match.}
    \label{table:main_results}
\end{center}
\end{figure*}

In this section, we consider the results of our method on the evaluation sets and experiments we conducted concerning the two stages of our method. The setup which produced the best results is discussed in \ref{sec:argument_labeling_exp}, including the chosen hyperparameters, which were tuned over the tuning sets.

\paragraph{Baseline} As a baseline for our approach, we considered the same process we used for generating the NomLex evaluation sets. More specifically, for a given parsed sentence with a given deverbal noun, our baseline method attempts to match the deverbal noun instance with all dependency patterns in appropriate entry within the adapted NomLex lexicon. Every fulfilled pattern should result in a set of labeled arguments. The combined set of non-colliding arguments, i.e., arguments that match a single argument type, are then mapped into pairs of headwords and UD relations, which are also the output of the baseline method.

\subsection{Argument Identification}
\label{sec:argument_identification_exp}

Using the set of relation labels in Section \ref{sec:argument_labeling} and considering each one of them as an argument candidate, we cover 94.6\% of all the relations in our paraphrasing-derived test-set, while producing 76 candidates (16.2\% of all proposed candidates) that are not arguments. We find this to be of sufficient coverage and accuracy for the paraphrasing dataset. Regarding the NomLex evaluation sets, all arguments are identified using that relations set (100\% coverage), while producing 24.8\% and 23.1\% non-argument candidates for $\text{NomLex}_{paraphrasing}$ and $\text{NomLex}_{other}$, respectively. As NomLex does not consider adjectival arguments, we choose to consider a reduced set of dependency relations without the \emph{amod} relation, keeping the same coverage and producing only 8.8\% and 8.7\% non-argument candidates, respectively. %

For the paraphrasing-derived dataset we also consider two other alternatives: relying on the information in the NomLex lexicon for each noun, resulting in coverage of 58.5\% and producing 6.9\% non-argument candidates, and relying on NomLex lexicon while also considering \emph{amod} relations, resulting in an increased coverage (85.3\%) and increased non-argument candidates (13.9\%). These low coverage results are anticipated as NomLex lexicon lacks the representation of some nominal structures, hence we chose the label-set approach as it was the most effective one.

We explored the resulted argument candidates and gathered three main reasons for the non-argument candidates. First, some correspond to arguments missing in the evaluation set. In the paraphrasing set, this is due to the focus on two arguments structure for each deverbal noun; In contrast, in the NomLex evaluation sets, this is primarily due to discarding of undetermined arguments and for the lack of prepositional adjuncts representation (which are captured using the dependency relations). Other resulted non-argument candidates are misaligned with the correct arguments, not sharing the same head-word, as emerged from a human-based evaluation set (such as paraphrasing-derived). Finally, the remaining non-arguments are indeed not an argument of the noun.

\subsection{Argument Labeling}

\label{sec:argument_labeling_exp}

\paragraph{Main Results} We experiment with two different labeling methods, as discussed in Section \ref{sec:argument_labeling}: nearest average of reference argument representations for each argument (nearest-avg-argument); k-nearest reference arguments (k-nearest-argument).
The results of the various labeling methods are shown in Table \ref{table:main_results} while considering the most suitable identification method for every evaluation set as raised from the argument identification comparison. We report our results on the three test sets and in comparison with the performance of the baseline method and naive `all-subject' and `all-object' methods (which label all argument relations with \emph{nsubj} and \emph{dobj}, respectively). 
As articulated from our results, both labeling methods performed better than the baseline regarding the paraphrasing evaluation set. Moreover, k-nearest-argument outperformed nearest-avg-argument on all metrics of all evaluation sets.
Best results were attained by calibrating the methods on the matching tuning sets, e.g., selecting a specific threshold for labeling $\emptyset$-typed arguments (0.56 for paraphrasing tune-set and 0.48 for NomLex tune-set). Yet, we examined similar performance tendencies between the tuning sets and the test sets (see Appendix \ref{appendix:argument_type_based_evaluation}), implying a generalization of our method for other examples. We further validated our method generalization for any arbitrary verb, by scoring relatively similar results on $\text{NomLex}_{other}$ as on $\text{NomLex}_{paraphrasing}$ without additional tuning, while each considers nouns that match a different set of verbs. The extended results in Appendix \ref{appendix:argument_type_based_evaluation} also demonstrate the Relation-F1 scores of our best method regarding the most common relations in the test sets.

\paragraph{Importance of Contextualization} Arguments of verbs and deverbal nouns share semantics, as both commonly paraphrase the same entity in different contexts. For instance, the subject of the verb \asstring{acquire} usually matches the semantic role of a `HUMAN' (\asstring{John acquired the ingredients}) or a `COMPANY' (\asstring{Apple acquired another startup company}). The same subjects can be realized in a deverbal noun context, as in \asstring{The ingredients acquisition of john} and \asstring{Apple's acquisition of the startup company}, correspondingly. The semantic role of words can be represented by vector representations, both contextualized representations as BERT and uncontextualized representations as Word2Vec \citep{mikolov2013efficient} vectors. We compared our main results with pre-trained BERT-based representations to uncontextualized representations, using pre-trained Fasttext Word2Vec model made by \citet{bojanowski2017enriching}. The results of our method regarding the two representations are shown in Table \ref{table:bert_vs_word2vec}. Using Word2Vec we see a decrease of about 25\% in Relation-F1 and about 40\% in Exact-Match compared to BERT results using our best method, from which we conclude that the context of the argument also affects the performance of our method.

\begin{figure}[hbt!]
\begin{center}    
    \begin{tabular}{l c c}
    \hline
    \multicolumn{1}{c}{Method} & BERT & Word2Vec \\
    \hline\hline
    nearest-avg-arg        & \textbf{44.08 (17.74)} & 20.78 (4.44) \\
    k-nearest-arg          & \textbf{62.93 (36.29)} & 46.53 (21.37) \\
    \hline
    \end{tabular}

    \captionof{table}{The best results of the suggested labelers using BERT and Word2Vec representations, on the paraphrasing test set, specified as ``Relation-F1 (Exact-Match)''.}
    \label{table:bert_vs_word2vec}
\end{center}
\end{figure}

\paragraph{Syntax vs Semantics} The previous experiment has demonstrated that the contextualized vectors outperform the static ones, suggesting the need for more than word semantics. In the following experiment, we further quantify the contribution of syntactic position vs. argument semantics to the final predictions. We manipulate the paraphrasing evaluation set by switching the sentence positions of the two specified arguments for each tagging sample. Note that the resulting sentence is usually neither grammatically nor semantically correct. Then, we apply our labeling stage while considering the BERT vectors over the arguments in the new positions. When compared to the labels of the same arguments received in the original positions, we see almost 70\% difference. Thus, the syntactic position has an innegligible effect on the verb-noun alignment that our method aims to resolve.

\subsection{Comparison to Earlier Work}
\label{sec:comparison_to_earlier_work}

Existing unsupervised attempts that approach the nominal argument labeling task as a transfer scenario from verbal arguments to nominal arguments (as our work), rely on a predefined semantic roles ontology. For instance, \citet{zhao2020unsupervised} consider SRL roles of verbs to label nouns with the same set of roles, as appears in CoNLL-2009 dataset \cite{hajivc2009conll}. Our defined task and proposed methods do not require a predefined semantic roles ontology, yet can be tested on one for comparability with such existing work. Thus, we apply our labeling methods on CoNLL-2009 nominal test data after verbalizing the nominal predicates in the dataset while considering the CoNLL-2009 verbal train data as verbal references. For evaluation comparability with \citet{zhao2020unsupervised}, we skip the argument identification stage and assume the identified arguments are given. Finally, we calculate the F1 performance (as discussed for ``Relation-F1'' in Section \ref{sec:EvaluationData}) of our methods, which we compare to the matching ones reported by \citet{zhao2020unsupervised}. As shown in Table \ref{table:conll_compare}, our best method (`k-nearest-argument') outperforms their baselines (`Most-frequent', `Factorization' and `Direct-transfer').
However, their `Full-system' approach transcends our method by exploiting a supervised verbal SRL system and data augmentations, which we do not use in our work.

\begin{figure}[hbt!]
\begin{center}    
    \begin{tabular}{l c}
    \hline
    \multicolumn{1}{c}{Method} & F1 \\
    \hline\hline
    Most-frequent                    & 56.51 \\
    Factorization                    & 44.48 \\
    Direct-transfer                  & 55.85 \\
    Full-system                      & \textbf{63.09} \\
    \hline
    k-nearest-argument (Ours)        & 58.82 \\
    \hline
    \end{tabular}

    \captionof{table}{F1 results reported by \citet{zhao2020unsupervised} on CoNLL-2009 nominal test data, compared to the result of our best labeler applied on the same dataset.}
    \label{table:conll_compare}
\end{center}
\end{figure}

%% file: 8_conclusions.tex
In this work, we formulate the task of aligning arguments of deverbal nouns to the arguments of their corresponding active verbal form. We formulate the task as a UD enrichment task, aiming to enrich deverbal nouns in text with verbal UD relations for the matching nominal arguments. Our formulation, compared to the ones suggested in previous works, does not rely on a predefined roles ontology.%

We suggest an unsupervised approach to this nominal-to-verbal argument mapping based on pre-trained contextualized word representations. Our method tries to match nominal identified arguments with automatically extracted arguments of the corresponding verb. The suggested method outperforms the NomLex-based baseline, which is based on an expertly constructed comprehensive lexicon. We also show the importance of contextualization, experiencing a 25\% decrease in performance when using uncontextualized vectors. Moreover, we further validate our hypothesis that semantics and syntactic structure are captured in the considered word representations using a dedicated experiment.

We provide a standalone code for enriching universal dependency trees with nominal arguments for a given parsed corpus, which can be integrated into NLP systems that use universal dependency patterns as part of their design or features.%

%% file: limitations.tex
The main drawback of the work is in its evaluation, which was performed on datasets which were not manually annotated for the task, but adapted to it in various means. While we believe these evaluation sets do provide a strong indication regarding task performance, evaluating on bespoke data explicitly annotated for the task is usually preferable. Another limitation is language specificity: the work currently focuses on English, without considering other languages, which are also left for future work.

%% file: appendix.tex
\section{Verb-Noun Argument Similarity}
\label{appendix:verb_noun_argument_similarity}

\begin{figure}[hbt!]
    \centering
    \begin{subfigure}[t]{0.83\columnwidth}
      \centering
      \includegraphics[width=\linewidth]{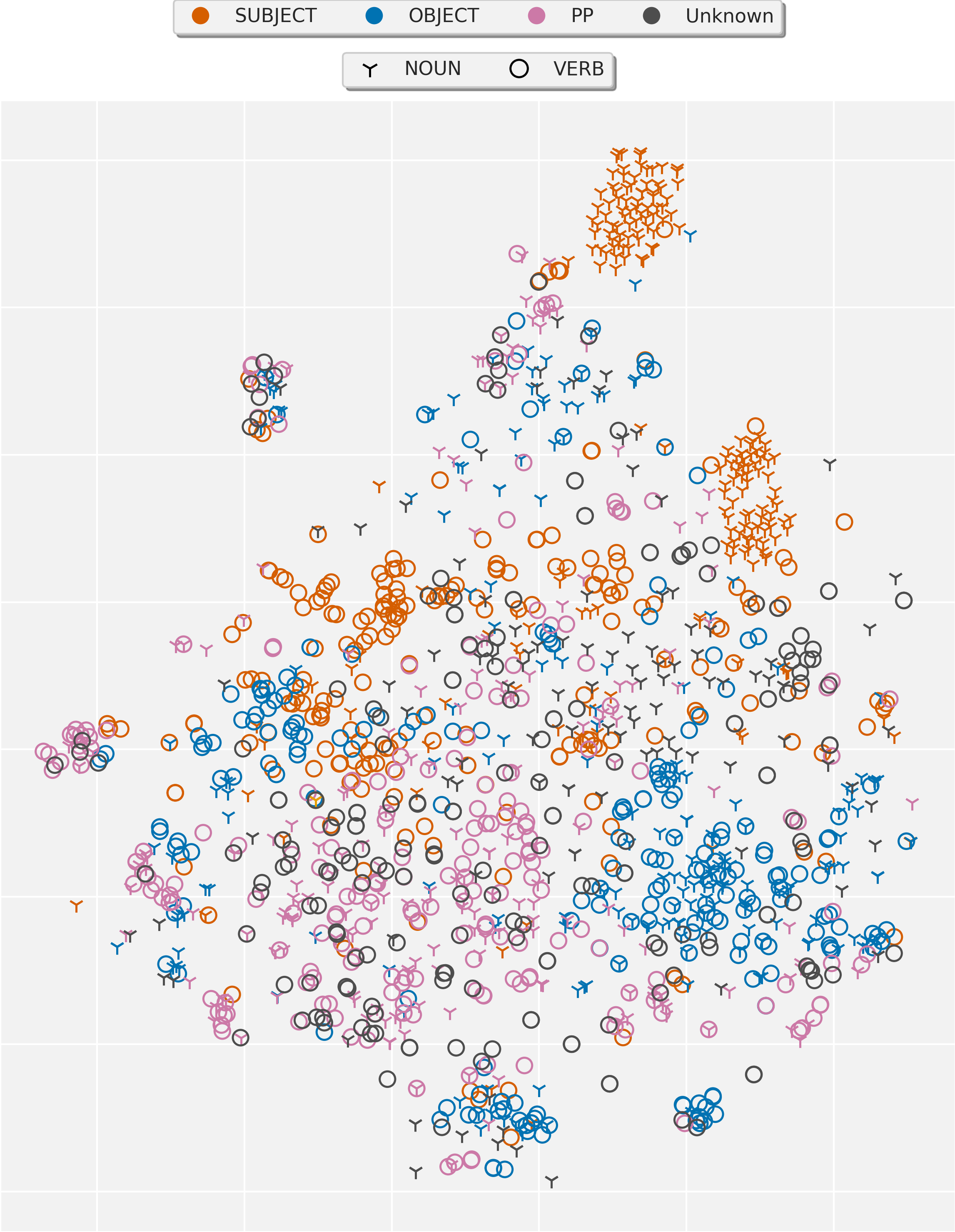}
      \caption{\asstring{transport}}
      \label{fig:arguments_repr_examine}
      \vspace{0.2cm}
    \end{subfigure}
    \begin{subfigure}[t]{0.83\columnwidth}
      \centering
      \includegraphics[width=\linewidth]{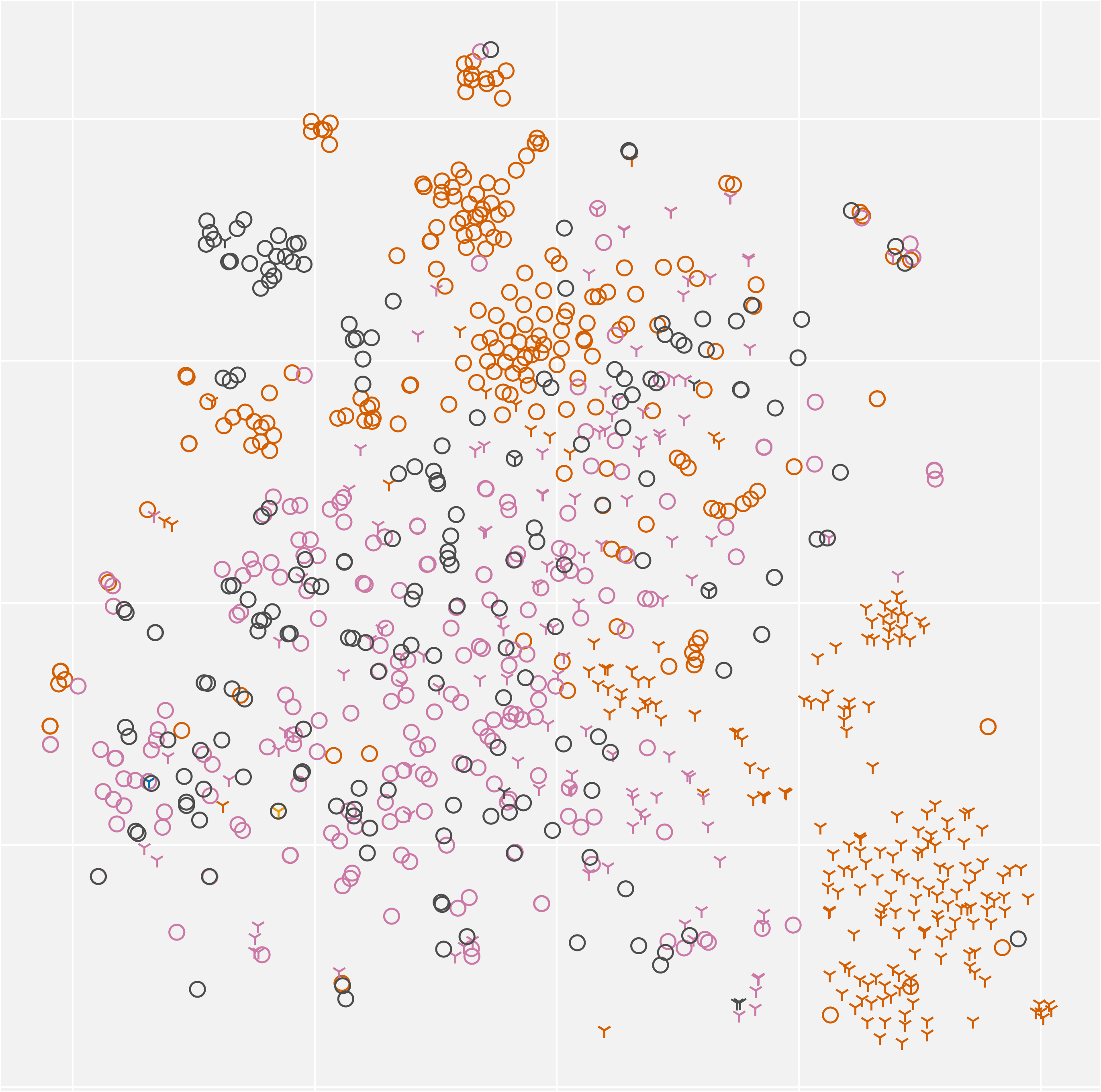}
      \caption{\asstring{participate}}
      \label{fig:arguments_repr_participate}
        \vspace{0.2cm}
    \end{subfigure}
    \begin{subfigure}[t]{0.83\columnwidth}
      \centering
      \includegraphics[width=\linewidth]{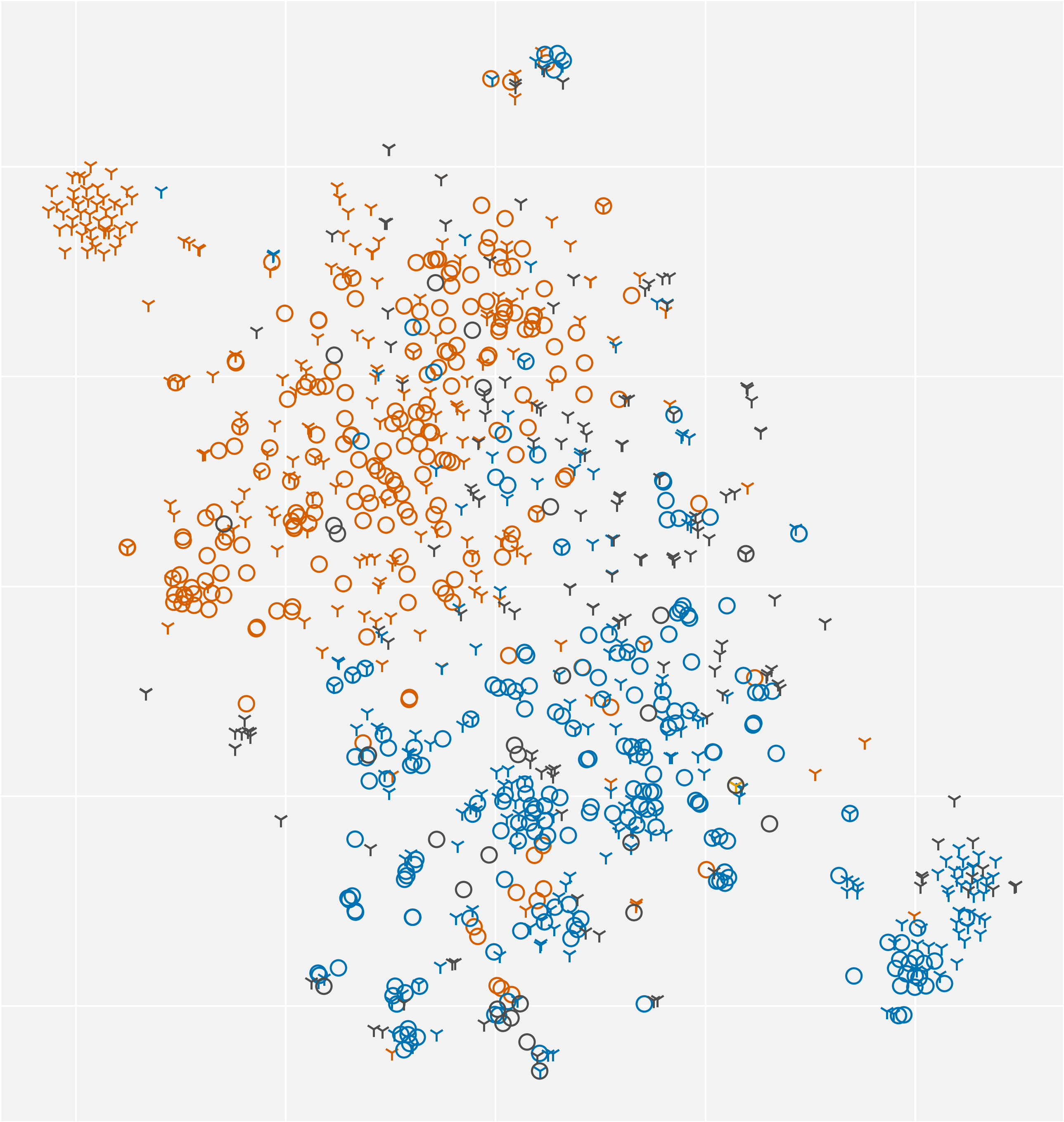}
      \caption{\asstring{violate}}
      \label{fig:arguments_repr_violate}
    \end{subfigure}
    
    \caption{Arguments representations of the verbs \asstring{transport}, \asstring{participate} and \asstring{violate} and their matching nouns}
    \label{fig:arguments_similarity_example}
\end{figure}

The similarity between arguments of verbs and arguments of matching deverbal noun realizations is a prominent requirement of our method. Similarly, \citet{zhao2020unsupervised} exploit verb-noun similarities and base their approach on this assumption. To explore this similarity, we take the verbal and nominal arguments extracted by NomLex of the types SUBJECT, OBJECT, PP, and undetermined (Unknown), embed them using a pre-trained BERT-large-uncased model, and compare their 2-dimensional representations (using t-SNE algorithm \cite{van2008visualizing} for dimension reduction). These representations are illustrated in Figure \ref{fig:arguments_similarity_example}, demonstrating relatively similar representations between arguments of the verbs \asstring{transport}, \asstring{participate} and \asstring{violate} (marked as 'O') and the matching arguments of the corresponding noun forms (marked as 'Y'). More concretely, most nominal argument representations in these illustrations have a nearby verbal argument neighbor with the correct argument type. This similarity establishes the foundation of our work.

\section{Extended Main Results}
\label{appendix:argument_type_based_evaluation}

We provide here more information regarding our best results. In Table \ref{table:main_results_dev}, we state the performance of all suggested methods when applied to the tuning sets, similar to our statement in Table \ref{table:main_results}.
Moreover, Table \ref{table:argument_type_based_scores} summarizes the number of instances for the most common verbal relations in each test set and the Relation-F1 score of every such relation. As expected, `\emph{nsubj}' and `\emph{dobj}' are the most common relations in the test sets. Other regarded relations are `\emph{nmod:x}' relations and $\emptyset$ relations (referring to non-argument identifications or predictions).

\begin{figure*}[hbt!]
\begin{center}
    \begin{tabular}{l c c c c c c}
    \hline
    \multicolumn{1}{c}{} & \multicolumn{2}{c}{Paraphrasing-derived} & \multicolumn{2}{c}{$\text{NomLex}_{paraphrasing}$} \\
    \multicolumn{1}{c}{Method} & Relation-F1 & Exact-Match & Relation-F1 & Exact-Match \\
    \hline\hline
    baseline (NomLex-based)     & 42.46 & 11.48 & - & - \\
    all-subject                 & 31.62 & 0.00 & 39.16 & 0.00 \\
    all-object                  & 34.78 & 0.00 & 37.92 & 0.00 \\
    nearest-avg-argument        & 54.62 & 21.31 & 44.96 & 21.99 \\
    k-nearest-argument          & \textbf{67.21} & \textbf{40.98} & \textbf{58.16} & \textbf{41.84} \\
    \hline
    \end{tabular}

    \captionof{table}{The best results of the two suggested labelers on the two tuning sets, compared to the baseline process and the naive methods `all-subject' and `all-object'.}
    \label{table:main_results_dev}
\end{center}
\end{figure*}

\begin{figure*}[hbt!]
\begin{center}    
    \begin{tabular}{l c c c c c c}
    \hline
    \multicolumn{1}{c}{} & \multicolumn{2}{c}{Paraphrasing-derived} & \multicolumn{2}{c}{$\text{NomLex}_{paraphrasing}$} & \multicolumn{2}{c}{$\text{NomLex}_{other}$}\\
    \multicolumn{1}{c}{Relation Type} & Support & F1 & Support & F1 & Support & F1 \\
    \hline\hline
    nsubj               & 151 & 71.34        & 1910 & 62.86        & 6825 & 61.01 \\
    dobj                & 202 & 79.49        & 2075 & 63.08        & 6277 & 63.50 \\
    $\emptyset$         & 58 & 9.45          & 382 & 13.22         & 1191 & 16.09 \\
    nmod:to             & 24 & 50.00         & 162 & 22.11         & 419 & 14.11 \\
    nmod:with           & 14 & 19.35         & 77 & 23.92          & 404 & 34.60 \\
    nmod:for            & 11 & 37.04         & 105 & 29.12         & 322 & 30.36 \\
    nmod:from           & 2 & 0.00           & 86 & 28.28          & 276 & 35.06 \\
    nmod:in             & 41 & 56.52         & 233 & 36.90         & 263 & 12.34 \\
    nmod:as             & 8 & 7.41           & 99 & 33.90          & 220 & 39.05 \\
    nmod:on             & 5 & 20.00          & 49 & 21.65          & 218 & 36.70 \\
    nmod:into           & 2 & 33.33          & 26 & 14.46          & 114 & 35.90 \\
    nmod:against        & 1 & 0.00           & 25 & 36.00          & 96 & 52.57 \\
    nmod:over           & 0 & -              & 12 & 0.00           & 76 & 35.46 \\
    nmod:about          & 1 & 0.00           & 0 & -               & 43 & 37.21 \\
    nmod:at             & 4 & 18.18          & 22 & 4.26           & 33 & 10.66 \\
    nmod:of             & 4 & 0.00           & 14 & 0.00           & 23 & 5.13 \\
    nmod:towards        & 0 & -              & 0 & -               & 17 & 51.43 \\
    nmod:through        & 11 & 0.00          & 8 & 17.14           & 13 & 21.05 \\
    nmod:across         & 0 & -              & 2 & 40.00           & 9 & 26.09 \\
    nmod:due to         & 0 & -              & 2 & 22.22           & 7 & 6.45 \\
    nmod:between        & 2 & 0.00           & 1 & 0.00            & 6 & 0.00 \\
    nmod:among          & 0 & -              & 1 & 33.33           & 6 & 7.41 \\
    nmod:along          & 1 & 66.67          & 0 & -               & 5 & 0.00 \\
    nmod:by             & 8 & 0.00           & 2 & 0.00            & 2 & 0.00 \\
    \hline
    \end{tabular}

    \captionof{table}{The support of the most common verbal relations in the test sets, alongside their Relation-F1 score (as `F1`) of our best method (`k-nearest-argument').}
    \label{table:argument_type_based_scores}
\end{center}
\end{figure*}

%% file: main.bbl
\begin{thebibliography}{23}
\expandafter\ifx\csname natexlab\endcsname\relax\def\natexlab#1{#1}\fi

\bibitem[{Angeli et~al.(2015)Angeli, Premkumar, and
  Manning}]{angeli2015leveraging}
Gabor Angeli, Melvin Jose~Johnson Premkumar, and Christopher~D Manning. 2015.
\newblock Leveraging linguistic structure for open domain information
  extraction.
\newblock In \emph{Proceedings of the 53rd Annual Meeting of the Association
  for Computational Linguistics and the 7th International Joint Conference on
  Natural Language Processing (Volume 1: Long Papers)}, pages 344--354.

\bibitem[{Baker et~al.(1998)Baker, Fillmore, and Lowe}]{baker1998berkeley}
Collin~F Baker, Charles~J Fillmore, and John~B Lowe. 1998.
\newblock The berkeley framenet project.
\newblock In \emph{36th Annual Meeting of the Association for Computational
  Linguistics and 17th International Conference on Computational Linguistics,
  Volume 1}, pages 86--90.

\bibitem[{Bojanowski et~al.(2017)Bojanowski, Grave, Joulin, and
  Mikolov}]{bojanowski2017enriching}
Piotr Bojanowski, Edouard Grave, Armand Joulin, and Tomas Mikolov. 2017.
\newblock Enriching word vectors with subword information.
\newblock \emph{Transactions of the Association for Computational Linguistics},
  5:135--146.

\bibitem[{Devlin et~al.(2018)Devlin, Chang, Lee, and
  Toutanova}]{devlin2018bert}
Jacob Devlin, Ming{-}Wei Chang, Kenton Lee, and Kristina Toutanova. 2018.
\newblock \href {http://arxiv.org/abs/1810.04805} {{BERT:} pre-training of deep
  bidirectional transformers for language understanding}.
\newblock \emph{CoRR}, abs/1810.04805.

\bibitem[{Etzioni et~al.(2008)Etzioni, Banko, Soderland, and
  Weld}]{etzioni2008open}
Oren Etzioni, Michele Banko, Stephen Soderland, and Daniel~S Weld. 2008.
\newblock Open information extraction from the web.
\newblock \emph{Communications of the ACM}, 51(12):68--74.

\bibitem[{Guo et~al.(2020)Guo, Dai, Vrande{\v{c}}i{\'c}, and
  Al-Rfou}]{guo2020wiki}
Mandy Guo, Zihang Dai, Denny Vrande{\v{c}}i{\'c}, and Rami Al-Rfou. 2020.
\newblock Wiki-40b: Multilingual language model dataset.
\newblock In \emph{Proceedings of The 12th Language Resources and Evaluation
  Conference}, pages 2440--2452.

\bibitem[{Gurevich et~al.(2007)Gurevich, Crouch, King, and
  De~Paiva}]{gurevich2007deverbal}
Olga Gurevich, Richard Crouch, Tracy~Holloway King, and Valeria De~Paiva. 2007.
\newblock \href
  {https://academic.oup.com/logcom/article-abstract/18/3/385/1746128} {Deverbal
  nouns in knowledge representation}.
\newblock \emph{Journal of Logic and Computation}, 18(3):385--404.

\bibitem[{Gurevich and Waterman(2009)}]{gurevich-waterman-2009-mining}
Olga Gurevich and Scott Waterman. 2009.
\newblock \href {https://www.aclweb.org/anthology/W09-2603} {Mining of parsed
  data to derive deverbal argument structure}.
\newblock In \emph{Proceedings of the 2009 Workshop on Grammar Engineering
  Across Frameworks ({GEAF} 2009)}, pages 19--27, Suntec, Singapore.
  Association for Computational Linguistics.

\bibitem[{Haji{\v{c}} et~al.(2009)Haji{\v{c}}, Ciaramita, Johansson, Kawahara,
  Mart{\'\i}, M{\`a}rquez, Meyers, Nivre, Pad{\'o}, {\v{S}}tep{\'a}nek
  et~al.}]{hajivc2009conll}
Jan Haji{\v{c}}, Massimiliano Ciaramita, Richard Johansson, Daisuke Kawahara,
  Maria~Ant{\`o}nia Mart{\'\i}, Llu{\'\i}s M{\`a}rquez, Adam Meyers, Joakim
  Nivre, Sebastian Pad{\'o}, Jan {\v{S}}tep{\'a}nek, et~al. 2009.
\newblock The conll-2009 shared task: Syntactic and semantic dependencies in
  multiple languages.

\bibitem[{Kilicoglu et~al.(2010)Kilicoglu, Fiszman, Rosemblat, Marimpietri, and
  Rindflesch}]{kilicoglu2010arguments}
Halil Kilicoglu, Marcelo Fiszman, Graciela Rosemblat, Sean Marimpietri, and
  Thomas~C Rindflesch. 2010.
\newblock \href {https://www.aclweb.org/anthology/W10-1906.pdf} {Arguments of
  nominals in semantic interpretation of biomedical text}.
\newblock In \emph{Proceedings of the 2010 workshop on biomedical natural
  language processing}, pages 46--54. Association for Computational
  Linguistics.

\bibitem[{Lapata(2000)}]{lapata2000automatic}
Maria Lapata. 2000.
\newblock \href {https://www.aaai.org/Papers/AAAI/2000/AAAI00-110.pdf} {The
  automatic interpretation of nominalizations}.
\newblock In \emph{AAAI/IAAI}, pages 716--721.

\bibitem[{Lee et~al.(2021)Lee, Lim, and Webster}]{lee-etal-2021-paraphrasing}
John Lee, Ho~Hung Lim, and Carol Webster. 2021.
\newblock \href {https://doi.org/10.18653/v1/2021.emnlp-main.632} {Paraphrasing
  compound nominalizations}.
\newblock In \emph{Proceedings of the 2021 Conference on Empirical Methods in
  Natural Language Processing}, pages 8023--8028, Online and Punta Cana,
  Dominican Republic. Association for Computational Linguistics.

\bibitem[{Macleod et~al.(1998)Macleod, Grishman, Meyers, Barrett, and
  Reeves}]{macleod1998nomlex}
Catherine Macleod, Ralph Grishman, Adam Meyers, Leslie Barrett, and Ruth
  Reeves. 1998.
\newblock \href
  {https://pdfs.semanticscholar.org/fc0e/3cf5f6d8bc457ec4989e9dfe1e5cd288c959.pdf}
  {Nomlex: A lexicon of nominalizations}.
\newblock In \emph{Proceedings of EURALEX}, volume~98, pages 187--193.

\bibitem[{Meyers et~al.(1998)Meyers, Macleod, Yangarber, Grishman, Barrett, and
  Reeves}]{meyers1998using}
Adam Meyers, Catherine Macleod, Roman Yangarber, Ralph Grishman, Leslie
  Barrett, and Ruth Reeves. 1998.
\newblock Using nomlex to produce nominalization patterns for information
  extraction.
\newblock In \emph{The Computational Treatment of Nominals}.

\bibitem[{Mikolov et~al.(2013)Mikolov, Chen, Corrado, and
  Dean}]{mikolov2013efficient}
Tomas Mikolov, Kai Chen, Greg Corrado, and Jeffrey Dean. 2013.
\newblock Efficient estimation of word representations in vector space.
\newblock \emph{arXiv preprint arXiv:1301.3781}.

\bibitem[{Nivre et~al.(2016)Nivre, De~Marneffe, Ginter, Goldberg, Hajic,
  Manning, McDonald, Petrov, Pyysalo, Silveira et~al.}]{nivre2016universal}
Joakim Nivre, Marie-Catherine De~Marneffe, Filip Ginter, Yoav Goldberg, Jan
  Hajic, Christopher~D Manning, Ryan McDonald, Slav Petrov, Sampo Pyysalo,
  Natalia Silveira, et~al. 2016.
\newblock Universal dependencies v1: A multilingual treebank collection.
\newblock In \emph{Proceedings of the Tenth International Conference on
  Language Resources and Evaluation (LREC'16)}, pages 1659--1666.

\bibitem[{Pad{\'o} et~al.(2008)Pad{\'o}, Pennacchiotti, and
  Sporleder}]{pado2008semantic}
Sebastian Pad{\'o}, Marco Pennacchiotti, and Caroline Sporleder. 2008.
\newblock \href
  {http://delivery.acm.org/10.1145/1600000/1599165/p665-pado.pdf?ip=176.231.62.115&id=1599165&acc=OPEN&key=4D4702B0C3E38B35\%2E4D4702B0C3E38B35\%2E4D4702B0C3E38B35\%2E6D218144511F3437&__acm__=1573170080_2b12da00cd99a89951467ee3732fd879}
  {Semantic role assignment for event nominalisations by leveraging verbal
  data}.
\newblock In \emph{Proceedings of the 22nd International Conference on
  Computational Linguistics-Volume 1}, pages 665--672. Association for
  Computational Linguistics.

\bibitem[{Pradhan et~al.(2004)Pradhan, Sun, Ward, Martin, and
  Jurafsky}]{pradhan2004parsing}
Sameer Pradhan, Honglin Sun, Wayne Ward, James~H Martin, and Dan Jurafsky.
  2004.
\newblock \href {https://nlp.stanford.edu/pubs/hlt-2004-noun.pdf} {Parsing
  arguments of nominalizations in english and chinese}.
\newblock In \emph{Proceedings of HLT-NAACL 2004: Short Papers}, pages
  141--144. Association for Computational Linguistics.

\bibitem[{Schuler(2005)}]{schuler2005verbnet}
Karin~Kipper Schuler. 2005.
\newblock \emph{VerbNet: A broad-coverage, comprehensive verb lexicon}.
\newblock University of Pennsylvania.

\bibitem[{Tiktinsky et~al.(2020)Tiktinsky, Goldberg, and
  Tsarfaty}]{tiktinsky2020pybart}
Aryeh Tiktinsky, Yoav Goldberg, and Reut Tsarfaty. 2020.
\newblock \href {http://arxiv.org/abs/2005.01306} {pybart: Evidence-based
  syntactic transformations for ie}.

\bibitem[{Van~der Maaten and Hinton(2008)}]{van2008visualizing}
Laurens Van~der Maaten and Geoffrey Hinton. 2008.
\newblock Visualizing data using t-sne.
\newblock \emph{Journal of machine learning research}, 9(11).

\bibitem[{Yahya et~al.(2014)Yahya, Whang, Gupta, and Halevy}]{yahya2014renoun}
Mohamed Yahya, Steven Whang, Rahul Gupta, and Alon Halevy. 2014.
\newblock Renoun: Fact extraction for nominal attributes.
\newblock In \emph{Proceedings of the 2014 conference on empirical methods in
  natural language processing (EMNLP)}, pages 325--335.

\bibitem[{Zhao and Titov(2020)}]{zhao2020unsupervised}
Yanpeng Zhao and Ivan Titov. 2020.
\newblock Unsupervised transfer of semantic role models from verbal to nominal
  domain.
\newblock \emph{arXiv preprint arXiv:2005.00278}.

\end{thebibliography}
